\documentclass{ws-us}
\usepackage[utf8]{inputenc}
\usepackage[sort,compress,super]{cite}
\usepackage{amsmath}
\usepackage{url}
\usepackage{dblfnote}
\usepackage{hyperref}
\usepackage{algorithm}
\usepackage{algpseudocode}
\usepackage{color}

\newcounter{Algorithm}

\counterwithout{algorithm}{section}

\begin{document}
\setlength{\pdfpagewidth}{8.5in}
\setlength{\pdfpageheight}{11in}
\catchline{0}{0}{2013}{}{}

\markboth{Zheng Yang et al.}{Paper Title}

\title{A Fast and Light-weight Non-Iterative \\ Visual Odometry with RGB-D Cameras}

\begin{NoHyper}
\author{Zheng Yang\footnotemark[1], Kuan Xu\footnotemark[2], Shenghai Yuan\footnotemark[3], Lihua Xie\footnotemark[4]}
\footnotetext{Email Addresses: *e220193@e.ntu.edu.sg, †kuan.xu@ntu.edu.sg, \\ ‡shyuan@ntu.edu.sg, §elhxie@ntu.edu.sg}
\address{School of Electrical and Electronic Engineering,\\
Nanyang Technological University, Singapore 639798, Singapore}
\maketitle
\end{NoHyper}

\begin{abstract}
In this paper, we introduce a novel approach for efficiently estimating the 6-Degree-of-Freedom (DoF) robot pose with a decoupled, non-iterative method that capitalizes on overlapping planar elements. Conventional RGB-D visual odometry(RGBD-VO) often relies on iterative optimization solvers to estimate pose and involves a process of feature extraction and matching. This results in significant computational burden and time delays. To address this, our innovative method for RGBD-VO separates the estimation of rotation and translation. Initially, we exploit the overlaid planar characteristics within the scene to calculate the rotation matrix. Following this, we utilize a kernel cross-correlator (KCC) to ascertain the translation. By sidestepping the resource-intensive iterative optimization and feature extraction and alignment procedures, our methodology offers improved computational efficacy, achieving a performance of 71Hz on a lower-end i5 CPU. When the RGBD-VO does not rely on feature points, our technique exhibits enhanced performance in low-texture degenerative environments compared to state-of-the-art methods.
\end{abstract}


\begin{multicols}{2}
\section{Introduction}

\subsection{Motivation}
Over the past several years, the domain of vision-based navigation and positioning systems has witnessed substantial growth in its application across a multitude of sectors. Notably, industries such as autonomous vehicles, unmanned aerial vehicles (UAVs), and virtual/augmented reality (VR/AR) have been increasingly incorporating Visual Odometry (VO) and Simultaneous Localization and Mapping (SLAM) for ego state estimations. 

The development of RGB-D cameras has notably improved visual odometry (VO) for indoor positioning, enabling a richer spatial analysis by providing both textural and geometric information. This sensor fusion enhances scene understanding and is pivotal for robust localization and mapping in VR/AR applications.

Most existing RGBD-VO algorithms [\citen{orb-slam2}, \citen{rtab}, \citen{3rgbd}, \citen{4rgbd}, \citen{rgbd-tam}] generally require feature point extraction and matching process which demands additional computations. All those methods require iterative algorithms to solve an optimization problem for estimating camera pose. However, such feature-based, iterative, optimization-centric approaches come with several drawbacks. Firstly, these feature-based methods heavily depend on the quality of feature points in texture-rich environments. When dealing with minimal texture environments, such as lab desks or walls in side standard rooms, the scarcity of feature points severely undermines the precision of feature point matching. Consequently, these methods often fail to deliver adequate accuracy in low-texture scenarios and may suffer from feature association due to a lack of unique feature points. Secondly, the processes of feature extraction, matching, and iterative optimization are typically computation-intensive, rendering many RGB-D solutions impractical for high-frame-rate activities such as playing VR/AR games.

To enhance the reliability of feature-based approaches, various techniques have attempted to leverage the structural data provided by RGB-D sensors, such as incorporating line and plane features into pose estimation [\citen{canny-vo}, \citen{edge-vo}, \citen{plane-edge-slam}]. However, these methods often demand more computations associated with additional feature extraction, matching, and iterative optimization. 

There are also works [\citen{OPVO}, \citen{LPVO}, \citen{Manhattanslam}] that introduce extra constraints to improve adaptability. For example, some approaches, as discussed in [\citen{Manhattanslam}], are based on the Manhattan-world hypothesis in a structured environment. This hypothesis assumes the presence of three mutually perpendicular planes in indoor spaces, forming a Manhattan coordinate system, which is ideal for AR/VR applications. However, they still rely on feature-based and iterative approaches that are computationally intensive. This is particularly challenging because AR/VR headsets often use low-performance chips, making these methods less suitable for such devices.

To address these issues, we propose a new non-iterative RGB-D visual odometry framework named Non-Iterative Depth-Enhanced Visual Odometry (NIDEVO). Unlike current solutions, we estimate the rotation and translation in a decoupled and non-iterative way, which makes our system very efficient. Besides, due to the independence from feature points and lines, our system achieves better performance than current methods in low-texture environments. In summary, our contributions include 
\begin{itemlist}
\item We design a non-iterative RGB-D visual odometry for AR/VR applications. The rotation and translation are estimated in a decoupled using a multi-threaded framework. Our system is very efficient and can run at a rate of 71Hz on a notebook PC.
\item We propose a novel non-iterative method for rotation estimation. This approach, based on the overlap assumption between two consecutive frames, allows for real-time estimation of 3-DoF rotation using only two sets of plane normal vectors without the need for complex plane fitting or vanishing point estimation steps.
\item We perform extensive experiments that prove the efficiency and effectiveness of the proposed methods. The results show that our approach outperforms state-of-the-art (SOTA) systems in low-texture scenes.
\end{itemlist}

\subsection{Related works}

\subsubsection{RGB-D Odometry}
In recent years, with the improvement of depth cameras, RGB-D camera-based visual odometry methods[\citen{orb-slam2},\citen{canny-vo},\citen{edge-vo},\citen{plane-edge-slam},\citen{Manhattanslam}, \citen{rgbd-dso}] have shown promising results. In the context of feature tracking pipelines in the VO study, the existing solutions can primarily be categorized into two methods: the direct method and the indirect method. The direct methods optimize the pose between two frames by minimizing the photometric error between the reference frame and the rectified current frame, for instance, [\citen{rgbd-dso}]. On the other hand, the indirect methods estimate the optimal pose by constructing the geometric error between matched features, for instance, [\citen{orb-slam2}]. However, in both cases, feature detection and tracking require a substantial amount of computational power.

RGBDTAM[\citen{rgbd-tam}] adopted a low-cost photometric error-based method for RGBD VO. Their cost function is a weighted sum of photometric error and inverse depth error. Unlike dense direct methods, this method does not compute the photometric error and inverse depth error for the whole image but instead focuses only on the points on the Canny edges. Therefore, RGBDTAM has a relatively lower computational cost and operational efficiency than dense direct methods. ORB-SLAM2[\citen{orb-slam2}] is a classic visual SLAM framework for localization methods using RGB-D cameras. It optimizes pose estimation by building 2D point-to-point errors from ORB features extracted from RGB images. 
To further enhance the robustness of feature-based methods, particularly in feature-sparse scenarios, 
the study [\citen{plane-edge-slam}] introduced an innovative method for estimating camera pose that relies on plane and edge information extracted from depth maps. This work meticulously extracts plane models and edges present in the scene and adds them to the optimization. 
However, its computational overhead is somewhat elevated due to additional plane fitting modules and optimization cost formulations when compared to conventional feature-point-based methods.


\subsubsection{Decoupled pose estimation method}
In addition to incorporating structured information into the optimization function for pose estimation, another class of methods decouples pose estimation into two steps: rotation and translation estimation[\citen{OPVO},\citen{LPVO},\citen{Manhattanslam},\citen{liu-manhattan}-\citen{linear-atlanta}]. These methods can directly use the plane and line information in the depth maps for pose estimation.

Some methods[\citen{OPVO},\citen{LPVO},\citen{Manhattanslam},\citen{liu-manhattan}] proposed the Manhattan assumption suitable for indoor scenarios. 
In [\citen{OPVO}], a method was proposed to construct the Manhattan coordinate system using plane normal vectors and then use visual feature points to construct a cost function for optimizing translation. 
However, the prerequisite for this approach is that at least three planes must be visible in each sequence frame, a requirement that is not always feasible in specific scenarios. To address this limitation, newer work [\citen{LPVO}] introduced a method that employs lines to establish a Manhattan coordinate system in scenes with fewer than three planes. 
[\citen{Manhattanslam}] offers the flexibility to adaptively employ the Manhattan-based method in scenes where Manhattan frames are present while resorting to point and line features for pose estimation in scenarios where the Manhattan assumption is not applicable. [\citen{liu-manhattan}] proposed a technique for estimating Manhattan coordinates using vanishing points. Despite their reliance on tracking the Manhattan coordinate system, often through the mean shift algorithm, these methods suffer from time-consuming clustering algorithms, impacting their real-time performance. Additionally, the Manhattan assumption imposes a strong constraint that limits its practical applicability. Although some studies [\citen{linear-atlanta}, \citen{atlanta}] have delved into the broader Atlanta assumption, these approaches also introduce their constraints, like the requirement for a minimum of two orthogonal planes and, occasionally, an IMU to determine gravity's direction.

Apart from methods based on the Manhattan assumption, some studies have achieved decoupled rotation and translation estimation with fewer assumptions, which divides the rotation estimation problem from iterative optimization, making the iterative optimization problem more tractable. In [\citen{vote}], a decoupled pose estimation method was proposed, which calculates rotation through the overlapped planes between two consecutive frames and subsequently obtains translation estimation using a general iterative solution. 
However, this method can only estimate rotation about the vertical axis and cannot estimate three degrees of freedom for rotation. Inspired by this approach, our method also uses point normal vectors to estimate rotation and combines the non-iterative solver from [\citen{ni-slam}] to complete a full 6 DOF pose estimation. Nevertheless, after calculating the point normal vectors, we do not directly compute many rotations. Instead, based on the overlap assumption, we select two sets of matching planes and compute the rotation. This is because at least two pairs of matching plane normal vectors are needed to determine a unique rotation [\citen{two-plane}].

\subsubsection{Organization of the paper}
The rest of this paper unfolds as follows: Section 2 introduces basic definitions and formulates the problem of decoupled rotation and translation estimation. In Section 3, we propose a non-iterative solution to this problem. Experimental results of our algorithm's implementation are presented in Section 4 and concluded in Section 5.

\section{Problem formulation}

\subsection{Basic definitions}
Before delving into the specifics of our proposed approach, we outline the formulation for both decoupled rotation and translation estimation. This subsection commences by presenting essential mathematical definitions and symbols that will be utilized in the ensuing discussions, followed by an overview of the foundational elements of problem formulation. In the subsequent subsection, the mechanics of the decoupling strategy for pose estimation will be elucidated.

Let us consider \( I_{r} \) to be the reference color image and \( I_{c} \) to be the current color image. Correspondingly, \( D_{r} \) serves as the reference depth image, while \( D_{c} \) functions as the current depth image. The normal map for the reference frame, derived from \( D_{r} \), is denoted by \( N_{r} \), and \( N_{c} \) is obtained from \( D_{c} \). We utilize \( \textbf{R}^{r}_{c} \) and \( \textbf{t}^{r}_{c} \) to signify the rotation matrix and the translation vector from the reference frame to the current frame, respectively,
\begin{equation}
\textbf{p}_r = \textbf{R}^r_c \cdot \textbf{p}_c + \textbf{t}^r_c\,,
\label{E1}
\end{equation}
where $\textbf{p}_{c}$ is a point in the coordinate of the current frame and $\textbf{p}_{r}$ is a point in the coordinate of the reference frame. 

For the component concerning rotation estimation, the issue of orientation correction is addressed through the utilization of a normal map. In this context, we employ \( \textbf{N}_{r/c}(u,v) \) to denote the local plane normal vector centered at the coordinate point \( (u,v) \) within the pixel coordinate system. Points that lie on the same plane tend to have similar local plane normal vectors. These akin vectors are aggregated into separate clusters, subsequently referred to as different \( Modes \). Each \( Modes[i] \) generally signifies a single global plane normal vector. Consequently, by leveraging these diverse global plane normal vectors, we can efficiently adjust the camera's orientation in a structured setting.

For the segment pertaining to translation estimation, we employ \( F(\cdot) \) to signify the Fast Fourier Transform, and \( \circ \) to denote element-wise multiplication. Unless explicitly stated in the text that follows, all other mathematical symbols are assumed to carry their standard mathematical interpretations.

\begin{figure*}
\begin{center}
\centerline{\includegraphics[width=\textwidth]{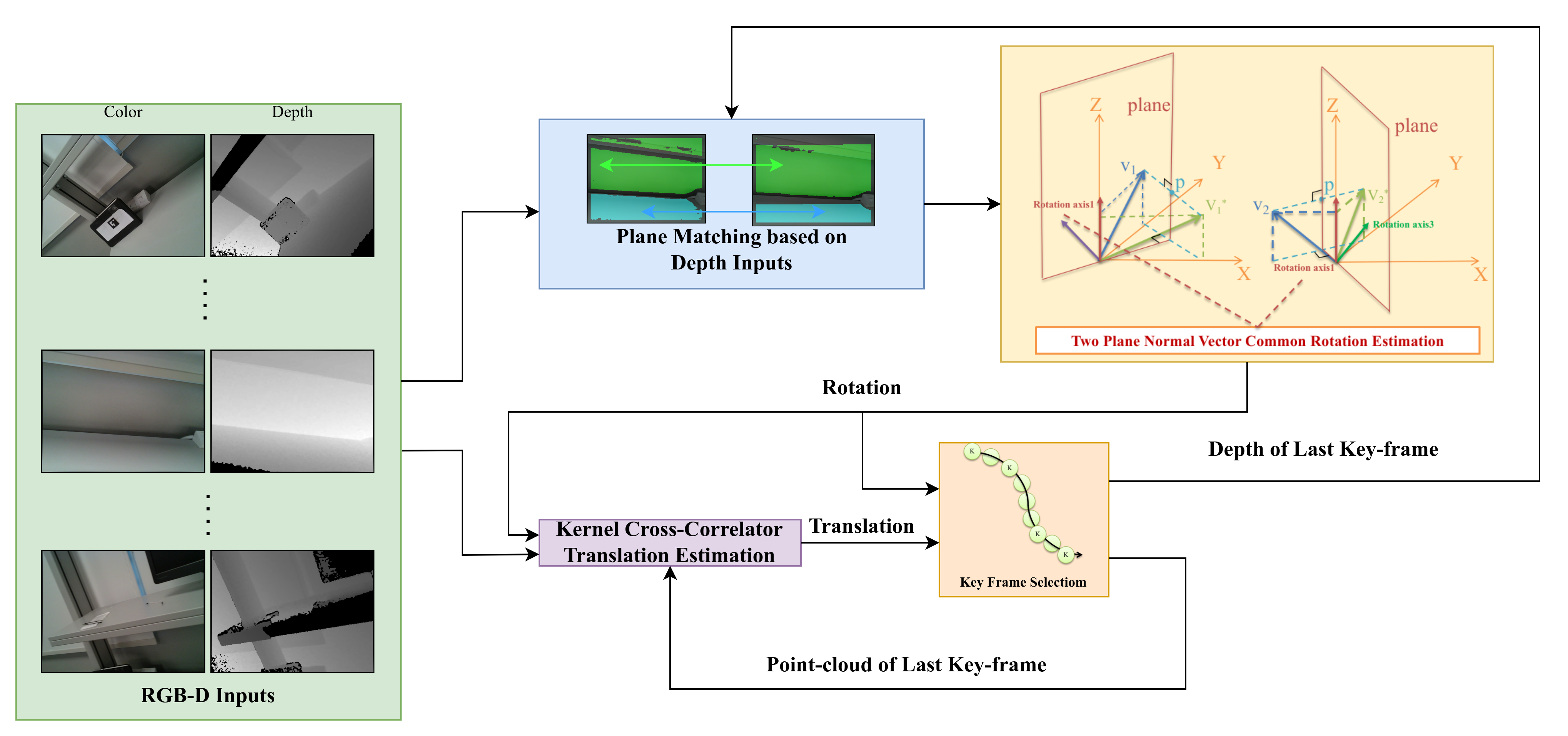}}
\caption{The rotation estimation in NIDEVO. It contains two sub-modules. The first one is the normal tracking part, which is introduced in Figure~\ref{plane_tracking_show}. The second one is the rotation calculation part, which relies on at least two planes.}
\label{rotation_estimation_algo}
\end{center}
\vspace{-0.4em}
\end{figure*}

\subsection{Decoupled rotation and translation estimation}
In this section, the decoupling of the 6-DOF pose estimation into a combination of the 3-DOF rotation estimation, 2-DOF planar translation estimation, and 1-DOF depth-directional shift calculation is elucidated. The conversion from the original 6-DoF pose estimation to a 3-DoF translation estimation is given by,
\begin{equation}
\textbf{p}^a_c = \textbf{R}^r_c \cdot \textbf{p}_c\,,
\label{E2}
\end{equation}
where $\textbf{p}^a_c$ represents the current point cloud whose orientation has been aligned to the reference frame. 

Following the description in [\citen{kcc}], kernel cross-correlator (KCC) is capable of rectifying the pose between two point cloud frames that have undergone rotation alignment. Therefore, before utilizing KCC, we first leverage the planar information within the scene to solve the 3-DOF rotation between the two point cloud frames. Under the premise that at least two overlapping planes exist between the successive point cloud frames, the rotational information between these planes can be directly used to represent the rotation between the two point cloud frames. Thus, we ascertain the 3-DOF rotation information between the two point cloud frames by matching the relationships of at least two pairs of planes across the frames. Following this, alignment can be performed between the two point clouds using (\ref{E2}).

To estimate 3-DoF translation, we use orthogonal projection to replace the respective projection for the aligned point cloud $\textbf{p}^a_c$ to obtain the orthogonal color and depth images as follows,
\begin{equation}
\begin{bmatrix}
    u \\
    v
\end{bmatrix}
= \frac{1}{r}
\begin{bmatrix}
    x^{a}_{c} \\
    y^{a}_{c}
\end{bmatrix}\,,
\label{E3}
\end{equation}
where $r$ represents the projection resolution, $[u, v]$ is the coordinate in the projection plane and $[x, y]$ is the coordinate in the aligned point cloud frame. The depth image is exactly the original depth map. The distinction between orthogonal projection and perspective projection lies in the fact that an orthogonally projected image can faithfully represent the actual size ratio between different objects, irrespective of their distance from the camera. In other words, orthogonal projection ensures that the same object or feature retains consistent size across two consecutive frames. Thus, by employing the orthogonally projected color image, the Kernel Cross-Correlator (KCC) can provide a reliable estimation of the two-degree-of-freedom (2-DOF) planar translation unaffected by the perspective effect. Subsequently, we can align the two depth images and use the mean difference of matched points between these images to represent the final degree of freedom (DOF) in translation estimation.

\section{Main Algorithm}
In this section, we first introduce the algorithm for the proposed depth-based rotation estimation. Subsequently, we present the translation estimation scheme in detail. Finally, we illustrate the design of the parallel program.

\subsection{Depth-based rotation estimation}
In visual odometry, a core issue in pose estimation is data association between two frames. As our approach employs a rotation matrix estimation method based on planar information, it is necessary to correctly associate planes within two frames to complete the rotation estimation.

Traditional normal vector estimation methods usually employ clustering or RANSAC to calculate the plane normal vectors in two-point clouds initially, then utilize the nearest neighbor algorithm to match similar plane normal vectors. Due to the use of clustering or RANSAC-based algorithms, these methods typically rely on high computational resources and have low computational efficiency. Conversely, we employ a straightforward yet potent approach to adeptly manage the issue of plane association between two depth maps. Figure~\ref{rotation_estimation_algo} illustrates the complete rotation estimation workflow.

Initially, drawing inspiration from [\citen{vote}], we compute a dense normal vector map from the depth image. Specifically, every image point represents the normal vector of the regional plane at that specific point's position, i.e.
\begin{equation}
\begin{split}
    \textbf{N}_r(u,v) = & (\textbf{p}_r(u,v+1)-\textbf{p}_r(u,v-1)) \\
    & \times (\textbf{p}_r(u-1,v)-\textbf{p}_r(u+1,v))\,.
\end{split}
\label{E4}
\end{equation}
Here, \( (u,v) \) denotes the pixel coordinates, \( \textbf{p}_{r} \) represents the point coordinates in the reference frame's camera coordinate system, and \( \times \) is the cross-product operator. Essentially, each point's normal vector is calculated from the local plane formed by its adjacent top, bottom, left, and right points. To reinforce the similarity of normal vectors on the same plane, we apply local smoothing via a sliding window to the precomputed normal vector map. This technique for normal vector map processing has been employed in previous studies [\citen{orb-slam2}, \citen{rtab}, \citen{3rgbd}], and its advantages are further validated in the experimental section as follows.

Employing this approach, we generate normal vector maps for both the reference and current frames aligned in the same pixel coordinate system. To circumvent the time-consuming nature of clustering algorithms, we employ a straightforward technique for associating distinct planes between two frames. The normal vectors at \( (u,v) \) in both the reference and current frames are represented as \( \textbf{N}_{r}(u,v) \) and \( \textbf{N}_{c}(u,v) \), respectively. A dot product nearing one between \( \textbf{N}_{r}(u,v) \) and \( \textbf{N}_{c}(u,v) \) implies that the point at \( (u,v) \) resides on the same plane in both the reference and current frame coordinate systems. 

Following this, we group overlapping points based on the resemblance of their normal vectors. Each such cluster essentially serves as a sub-representation of the normal vector for a specific plane. To derive a singular normal vector that represents a plane, we employ a straightforward yet robust technique: using the median of these sub-representations as the definitive plane normal. This choice is motivated by the median's greater resilience to outliers compared to the mean. The subsequent experimental section further validates the efficacy of using the median of normal vector sets as the global plane normal. While conceptually akin to density-based clustering, our approach is more efficient as it focuses solely on co-planar points rather than the entire point cloud. A clearer illustration of this concept is provided in Algorithm~\hyperref[alg:plane_tracking]{\ref*{alg:plane_tracking}}. Figure~\ref{plane_tracking_show} shows the effect of the plane tracking technique on the ICL-NUIM dataset[\citen{icl}]. 

After the identification of tracked planes in both the reference and current frames, our rotation estimation module comes into play. The sole prerequisite for our approach is the existence of at least two non-parallel, overlapping planes between the two frames, which is a condition generally met in indoor settings. In the most limiting scenario, assume that we have identified two sets of non-parallel planes via the preceding plane tracking algorithm. We designate their corresponding normal vectors as \( \textbf{N}^{P1}_{r} \), \( \textbf{N}^{P1}_{c} \), \( \textbf{N}^{P2}_{r} \), and \( \textbf{N}^{P2}_{c} \), respectively. Upon acquiring \( \textbf{N}^{P1}_{r} \), \( \textbf{N}^{P1}_{c} \), \( \textbf{N}^{P2}_{r} \), and \( \textbf{N}^{P2}_{c} \), we proceed with a unique rotation


\refstepcounter{Algorithm}
\label{alg:plane_tracking}

\begin{center}
\begin{tabular}{|p{0.95\linewidth}|}
\hline
\noindent\textbf{Algorithm \theAlgorithm:} Plane Tracking Algorithm\\
\hline
\begin{algorithmic}[1]
\State \textbf{Input:} Reference normal map
\State \textbf{Output:} Mode list (Modes)
\For{each point $(u,v)$ on reference normal map}
    \If{$N_{u,v} \times N_{u,v} \geq \text{Threshold}_0$}
        \For{each previous Mode in Modes}
            \If{$N_{u,v} \times \text{Mode}[i] \geq \text{Threshold}_1$}
                \State Mode\_Found $\gets$ True
                \State \textbf{exit loop}
            \EndIf
        \EndFor
        \If{not Mode\_Found}
            \State Add $N_{u,v}$ as a new Mode to Modes
        \EndIf
    \EndIf
\EndFor
\end{algorithmic} \\
\hline
\end{tabular}
\end{center}estimation, employing a technique akin to the one outlined in the reference [\citen{two-plane}]. Rotation estimation based on these two sets of plane normal vectors can be categorized into three distinct scenarios: 
\begin{itemlist}
\item The first situation is that $\textbf{N}^{P1}_{r}$ equals $\textbf{N}^{P1}_{c}$ and $\textbf{N}^{P2}_{r}$ equals $\textbf{N}^{P2}_{c}$. This indicates that the rotation between the two frames is 0, so the rotation matrix is an identity matrix.
\end{itemlist}
\begin{figurehere}
\begin{center}
\centerline{\includegraphics[width=3in]{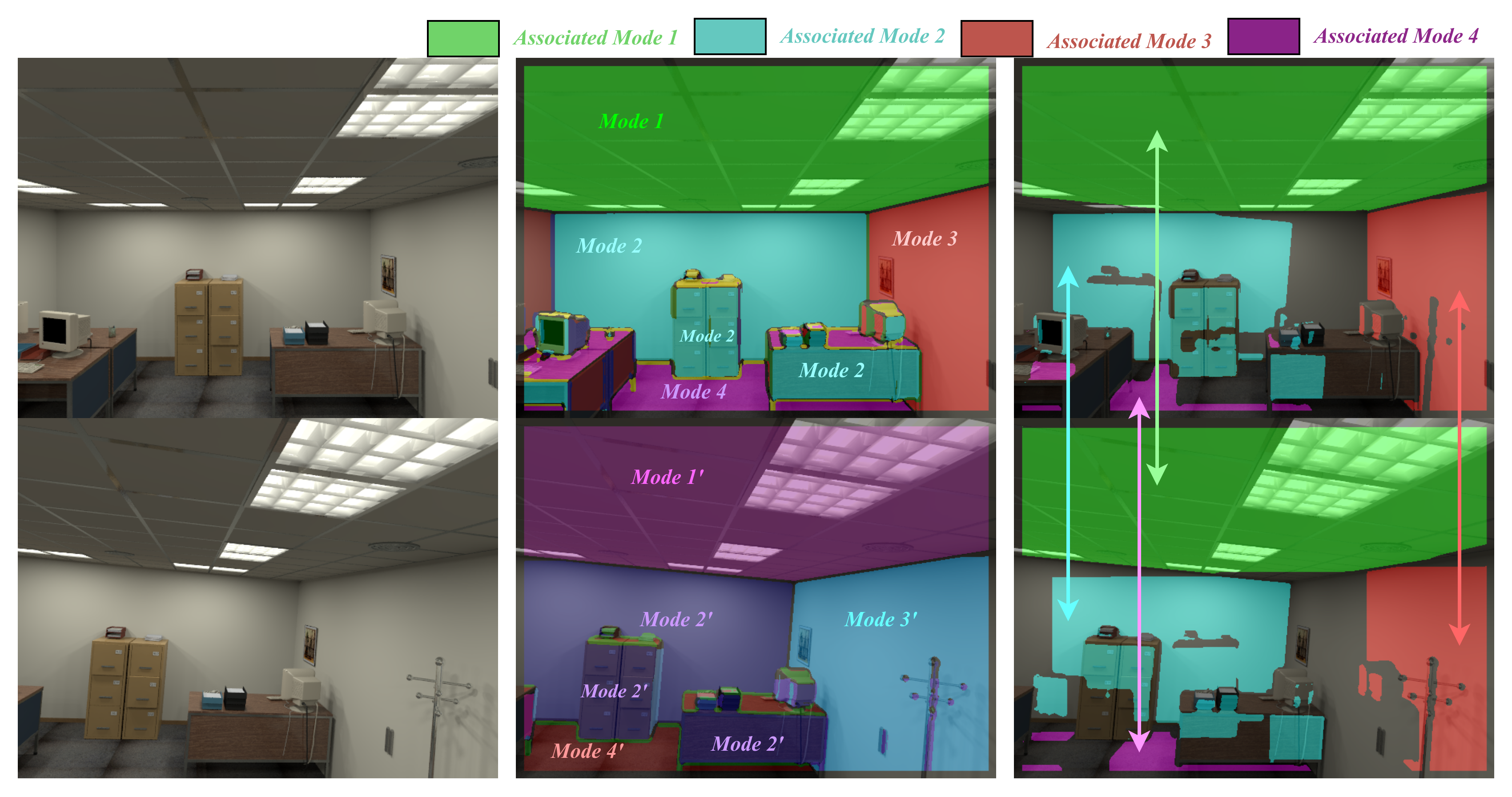}}
\caption{The plane tracking algorithm results on the ICL-NUIM dataset[\citen{icl}]. The first column depicts the original two frames, while the second column shows their plane modes, respectively. The final column shows the tracking effect of the proposed method. The pictures display the tracking results between two consecutive frames, with identical planes presented in matching colors.}
\label{plane_tracking_show}
\end{center}
\end{figurehere}
\begin{itemlist}
\item The second situation is that $\textbf{N}^{P1}_{r}$ equals $\textbf{N}^{P1}_{c}$ or $\textbf{N}^{P2}_{r}$ equals $\textbf{N}^{P2}_{c}$, which means that the rotation axis between the two frames is either $\textbf{N}^{P1}_{r}$ or $\textbf{N}^{P2}_{r}$. According to the Rodrigues' rotation formula
\begin{equation}
\textbf{N}_r=\cos\alpha \cdot \textbf{N}_c + (1-\cos\alpha) \cdot (\textbf{N} \cdot \textbf{N}_c) \cdot \textbf{N}+\sin\alpha \cdot \textbf{N} \times \textbf{N}_c\,, 
\label{E5}
\end{equation}
where $\textbf{N}$ represents the rotation axis between the two vectors, and $\alpha$ represents the rotation angle of the two vectors around this axis. It indicates that when the rotation axis between two vectors is known, the unique rotation angle around this axis can be calculated as follows:
\begin{equation}
\cos\alpha=\frac{\textbf{N}_r \cdot \textbf{N}_c - (\textbf{N} \cdot \textbf{N}_c)^2}{1 - (\textbf{N} \cdot \textbf{N}_c)^2}\,,
\label{E6}
\end{equation}
\begin{equation}
\sin\alpha=\frac{\textbf{N}_r \cdot (\textbf{N} \times \textbf{N}_c)}{\|\textbf{N} \times \textbf{N}_c\|^2}\,. 
\label{E7}
\end{equation}
\item In the third situation, which is the most general case, $\textbf{N}^{P1}_{r}$ is not equal to $\textbf{N}^{P1}_{c}$ and $\textbf{N}^{P2}_{r}$ is not equal to $\textbf{N}^{P2}_{c}$. In this situation, we cannot directly get the rotation axis between the two frames, but we know that the rotation axis must be orthogonal to $\textbf{N}^{P1}_{r}-\textbf{N}^{P1}_{c}$ and also to $\textbf{N}^{P2}_{r}-\textbf{N}^{P2}_{c}$, so we can calculate the rotation axis $\textbf{N}$ as follows:,
\begin{equation}
\textbf{N}=\frac{(\textbf{N}^{P1}_r-\textbf{N}^{P1}_c) \times (\textbf{N}^{P2}_r-\textbf{N}^{P2}_c)}{\|(\textbf{N}^{P1}_r-\textbf{N}^{P1}_c) \times (\textbf{N}^{P2}_r-\textbf{N}^{P2}_c)\|}\,. 
\label{E8}
\end{equation}
\end{itemlist}

It's important to highlight that [\citen{vote}] employs a plane tracking algorithm similar to ours. However, its rotation calculation diverges by computing a rotation for each pair of point normal vectors, subsequently using a voting mechanism to finalize the optimal rotation. This method is limited to estimating 1-DoF rotation, given that a unique rotation matrix can only be determined when the rotation axis is specified. In two-dimensional planar motion, this axis is invariably parallel to the vertical axis.

When more than two pairs of planes are associated, we compute the outcomes for all possible permutations and combinations. The remaining planes are then used to assess the error of the rotation matrix derived from each permutation and combination. Ultimately, the rotation matrix with the least error is selected as the final estimate. This approach allows us to leverage plane information in the scene for robust rotation estimation. 

\subsection{Kernel Cross-Correlator-based translation estimation}
As discussed in the preceding sections, we have already obtained the estimated rotation matrix, leaving only three degrees of freedom in position estimation $(x,y,z)$ to be determined. In alignment with the scheme of decoupled pose estimation detailed in Section 2, we will first estimate the planar translation and subsequently correct the depth-directional movement.

We adopt a similar pipeline to [\citen{ni-slam}] for translation estimation. Assume the orthogonal projected image size is $N \times M$. We can express the image as a column vector $\textbf{x}$, where the length $n$ of $\textbf{x}$ equals $N \times M$. Therefore, the planar shift of the image can be expressed as the circle shift of the column vector. In accordance with the principles of KCC[\citen{kcc}], the task of translation estimation can be reformulated as an optimization problem that possesses a closed-form analytical solution. So now we define training samples for the reference frame $\textbf{x}$, and obtain $\textbf{x}_{0}$, $\textbf{x}_{1}$, $\textbf{x}_{2}$, $...$, $\textbf{x}_{n-1}$. The subscript $i$ of $\textbf{x}$ represents the number of the circle shifts. At this point, we hope to construct a mapping relationship $y_{i} = f(\textbf{x}_{i})$ that can map $\textbf{x}_{i}$ from $R^{n}$ to a real number $y_{i}$. The magnitude of this real number represents the confidence of the circle shifting $i$ positions.

This kind of mapping definition is reasonable: when we get the column vector $\textbf{z}$ of the current frame, we can perform a circle shift on the current frame $\textbf{z}$, resulting in $\textbf{z}_{0}$, $\textbf{z}_{1}$, $\textbf{z}_{2}$, $...$, $\textbf{z}_{n-1}$. After we input these circles shifted $\textbf{z}_{i}$ into the previously well-trained mapping relationship $f(\cdot)$, there should be a $\textbf{z}_{i}$ corresponding to the largest $f(\textbf{z}_{i})$. If we let $y_{0}$ equal $1$ during training, while all other $y_{i}$ equal $0$, then the largest $y_{i}$ corresponds to $\textbf{z}_{i}$, which can be understood as the most similar $\textbf{z}_{i}$ to $\textbf{x}_{0}$. In other words, after being circle-shifted by $i$ positions, $\textbf{z}$ exhibits the highest similarity to $x$. Here, $\textbf{x}$ serves as the reference frame, while $\textbf{z}$ represents the current frame. Thus, this $i$th circle shift pattern signifies the planar transition from $\textbf{x}$ to $\textbf{z}$.

Therefore, the most critical step is how to train such a mapping function, $y_{i}=f(\textbf{x}_{i})=\textbf{b}^{T}\textbf{x}_{i}$ for $i=1,2,3,...,n-1$. This problem can be constructed as an optimization problem, and we can express the objective function as
\begin{equation}
\textbf{b}^{\ast}=\underset{b}{\mathrm{argmin}} \sum_{i=0}^{n-1}(\textbf{b}^T\textbf{x}_i-y_i)^2+\lambda\|\textbf{b}\|^2,
\label{E9}
\end{equation}
where $\lambda$ is the weight of the regularization term. This optimization problem can be solved by a closed-form solution by setting its first derivative to $0$. A closed-form solution (\ref{E10}) is shown in a complex domain:
\begin{equation}
\textbf{b}^{\ast}=(\textbf{X}^H\textbf{X}+\lambda \textbf{I})^{-1}\textbf{X}^H\textbf{y}\,, \label{E10}
\end{equation}
where $\textbf{X}$ denotes the circulant matrix of $\textbf{x}$, so that it can be generated by its first row $\textbf{x}^T$, and H is the conjugate transpose. Similar to [\citen{ni-slam}], to avoid the huge computation of the inverse matrix $(\textbf{X}^H\textbf{X}+\lambda \textbf{I})^{-1}$, we convert the solution to the frequency domain as (\ref{E11}):
\begin{equation}
F(\textbf{b}^{\ast})=\frac{F^{\ast}(\textbf{x}) \circ F(\textbf{y})}{F^{\ast}(\textbf{x}) \circ F(\textbf{x})+\lambda},
\label{E11}
\end{equation}
where $F(\cdot)$ is the discrete Fourier transform and the superscript operator $\ast$ is the complex conjugate, $\circ$ and $\div$ denote the element-wise multiplication and division respectively.

However, the above objective function can only achieve linear mapping and may not be suitable for our application scenario. Specifically, when observing the function, we can find that it is an objective function of a classification problem, where $y_i$ can be understood as the label of $\textbf{x}_i$. We hope to perform linear classification in the $\textbf{R}^n$ space of $\textbf{x}$. In classification problems, linear classification results are often unobtainable in the original feature space. However, non-linearly separable problems in low-dimensional space can often be linearly separable by mapping to high-dimensional space. This mapping from low-dimensional to high-dimensional space is assumed to be $\phi(\cdot)$. After mapping to high dimensional space, $\textbf{x}_i$ becomes $\phi(\textbf{x}_i)$, namely $\textbf{b}^{\ast}=\sum_{i=0}^{n-1}\alpha_i \phi(\textbf{x}_i)$. So $f(\textbf{z})=\sum_{i=0}^{n-1}\alpha_i\phi(\textbf{x}_i)\phi({\textbf{z}})$. After introducing the kernel function, $f(\textbf{z})=\sum_{i=0}^{n-1}\alpha_i\phi(\textbf{x}_i)\phi({\textbf{z}})=\sum_{i=0}^{n-1}\alpha_i k(\textbf{x}_i, \textbf{z})$. In this paper, we select the Gaussian kernel as our kernel function.

After transforming into the high-dimensional space, the original problem is transformed into finding a set of optimal $\alpha_i$. The solution can be given (\ref{E12}):
\begin{equation}
\boldsymbol{\alpha}=(\textbf{K}+\lambda \textbf{I})^{-1}\textbf{y},
\label{E12}
\end{equation}
where $\boldsymbol{\alpha}=[\alpha_0, \alpha_1, ..., \alpha_{n-1}]^T$, $\textbf{K}$ is the kernel matrix with each element $k_{i,j}=k(\textbf{x}_i, \textbf{x}_j)$. Then (\ref{E12}) can be calculated in the frequency domain:
\begin{equation}
F(\boldsymbol{\alpha})=\frac{F(\textbf{y})}{F(\boldsymbol{k^{xx}})+\lambda},
\label{E13}
\end{equation}
where $\boldsymbol{k^{xx}}$ is the first row of the kernel matrix $\textbf{K}$. To be more robust, all the circular shifts of a sample $\textbf{z}$ are tested. Define the kernel matrix $\boldsymbol{K^{zx}}$ where each element $k_{i,j}=k(\textbf{z}_i, \textbf{x}_j)$ and $\textbf{z}_i$ is the $i$th row of the circulant matrix $\textbf{Z}$. Then we can derive (\ref{E14}):
\begin{equation}
F(\textbf{z})=K^{\textbf{zx}}\boldsymbol{\alpha},
\label{E14}
\end{equation}
given that $f(\textbf{z})=[f(\textbf{z}_0), f(\textbf{z}1), ..., f(\textbf{z}{n-1})]^T$, the reactions of the samples that underwent circular shifts are identifiable in the frequency domain:
\begin{equation}
F(f(\textbf{z}))=F(\boldsymbol{k^{\textbf{z}\textbf{x}}}) \circ F(\boldsymbol{\alpha}).
\label{E15}
\end{equation}
When calculating the planar displacement, we only need to select the largest $f(\textbf{z}_i)$ corresponding to $i$ as the shift of the column vector $\textbf{z}$, and then reshape to the original image size. If we assume the estimated translation to be $(\delta i, \delta j)$ with pixel unit, the inferred translation on the axonometric plane, represented by $(\delta x, \delta y)$, can be expressed as an element-wise multiplication:
\begin{equation}
(\delta x, \delta y)=(r_x, r_y) \circ (\delta i, \delta j),
\label{E16}
\end{equation}
where $r_x$ and $r_y$ are the image resolutions in $x$ and $y$ directions respectively.

As the viewpoint of the camera alters, the intersection between the current and
keyframes diminishes, leading to a feeble peak intensity. This evaluation of peak intensity is denoted as the Peak to Sidelobe Ratio (PSR). In [\citen{ni-slam}], they proposed the PSR as 
\begin{equation}
PSR=\frac{\max F_{i,j}(\textbf{z})-\mu_s}{\theta_s},
\label{E17}
\end{equation}
where $\mu_s$ represents the mean of the sidelobe and $\theta_s$ denotes its standard deviation.

The PSR is a measure of similarity between two sets of point clouds and is perceived as an activator to incorporate a new keyframe into the map. Finally, for the depth-directional translation estimation, we use Equation~\ref{E18}, which averages the differences between the well-matched pixels, 
\begin{equation}
\delta z=ave(s_{\delta i, \delta j}(\textbf{I}^d_{i,j})-\textbf{I}^{kd}_{i,j}),
\label{E18}
\end{equation}
where $(i,j) \in \{(i,j)|\rho(s_{\delta i, \delta j}(\textbf{I}^{c}_{i,j})-\textbf{I}^{kc}_{i,j})<\textbf{T}_c\}$ and $\rho(\cdot)$ is a general objective function ($L_1$-norm in the tests). $(\delta i, \delta j)$ is the estimated image translation in Equation~\ref{E16} and $s_{\delta i, \delta j}(\cdot)$ is the shift of an image by $(\delta i, \delta j)$ pixels. $I^{d}$, $I^{kd}$, $I^{c}$, and $I^{kc}$ are the depth, key depth, color, and key color images, respectively. Therefore, the determined displacement, represented as $\boldsymbol{\delta p} = [\delta x, \delta y, \delta z]^T$, originates from the isolated translation in the axonometric plane combined with the depth orientation.

\subsection{The combination of decoupled rotation and translation estimation} 
Figure~\ref{architecture} shows the overall architecture of the proposed RGB-D odometry. Because we adopt a decoupling approach in this paper to estimate rotation and translation separately, we can optimize the efficiency of our program through multithreading. When the normal map calculation thread receives a newly read depth map, it will calculate the normal map and push the result in buffer1. Within Figure~\ref{architecture}, the dashed lines represent the logical flow of the system, while the solid lines depict the actual flow of the system after parallel programming.
\begin{figurehere}
\begin{center}
\centerline{\includegraphics[width=3in]{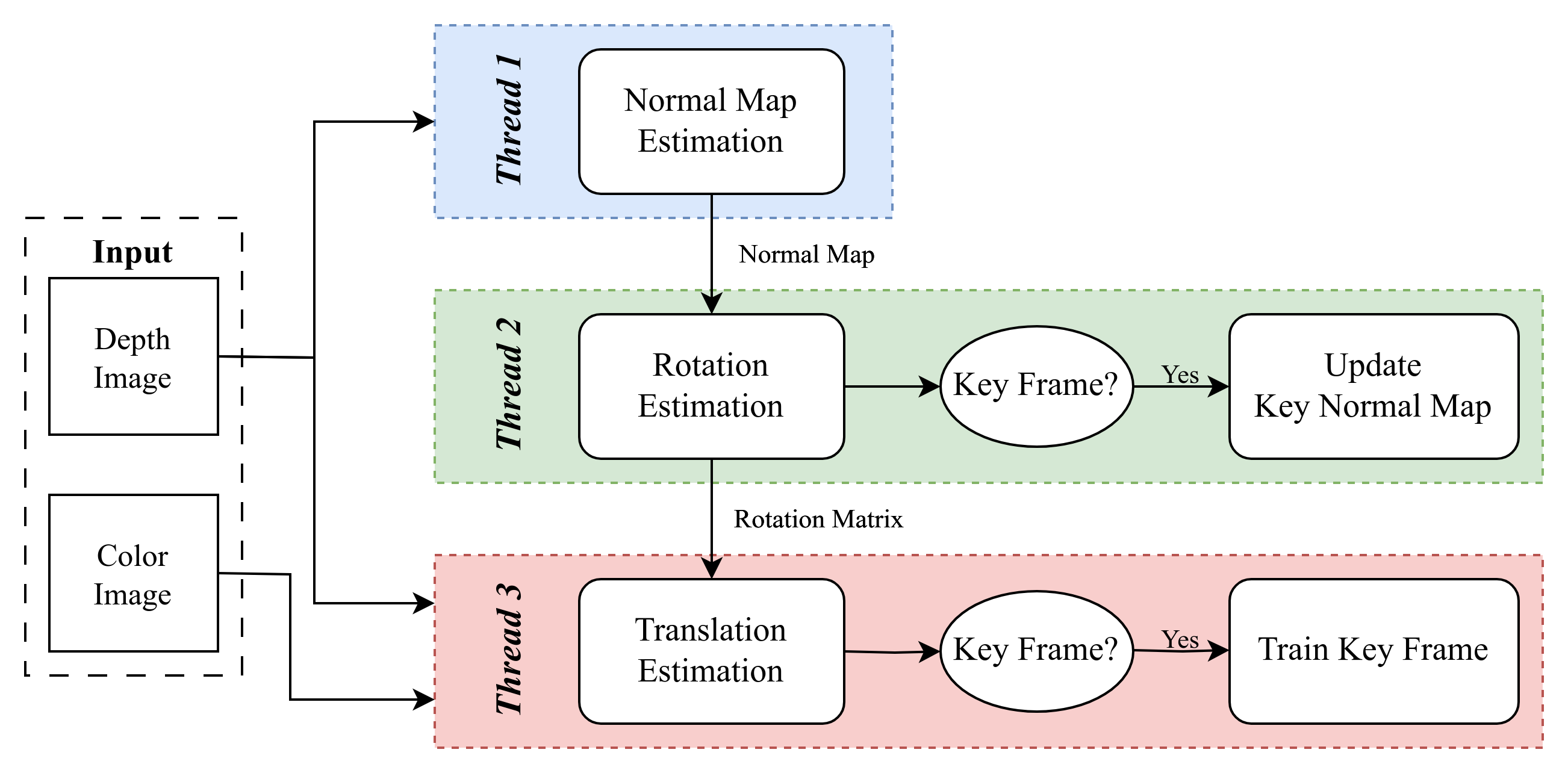}}
\caption{The proposed paralleled architecture with three threads for normal vector, rotation, and translation estimation. Keyframe selection strategies are independently applied in rotation and translation threads.}
\label{architecture}
\end{center}
\end{figurehere}
We divide the program into three threads: one thread for processing the calculation of the normal map, one thread for rotation estimation, and another thread for displacement estimation. This parallel design allows each program to run independently without waiting for each other. 

It is important to note that the running efficiency of the parallel version is limited by the slowest thread in terms of single-frame processing time, which is the rotation estimation thread in this case.

\section{Experimental Results}
In this section, we present experimental results on ICL-NUIM datasets[\citen{icl}] and self-collected datasets. We use ORB-SLAM2[\citen{orb-slam2}], RTAB-MAP[\citen{rtab}] and some state-of-the-art decoupling methods[\citen{OPVO}, \citen{LPVO}] as baselines for comparison.

\subsection{Evaluation method}
For ICL-NUIM datasets, we evaluate our estimated trajectories with the ground-truth trajectories by using the Root Mean Square Error (RMSE) metric over the aligned trajectories. Denote the difference between ground truth and estimated position on $x, y, z$-axis as $\delta x_i, \delta y_i, \delta z_i$, respectively. Then, the RMSE\begin{table*}
\small
\tbl{The comparison RMSE (m) results on ICL-NUIM Datasets[\citen{icl}]. The best results are highlighted in \textbf{bold}, and the second most accurate results are \underline{underlined}.\\ \label{tab2}}
{\resizebox{7.5in}{!}{
\begin{tabular}{@{}lcccccccr@{}}
\toprule
Datasets & ORB-SLAM2 & ORB-RTAB & BRISK-RTAB & KAZE-RTAB & OPVO[\citen{OPVO}] & LPVO[\citen{LPVO}] & NIDEVO (Ours) \\ \colrule
office seq0 & \underline{0.0240} & 0.2319 & 0.2333 & 0.3302 & 0.0480 & 0.0600 & \textbf{0.0220} \\
office seq1 & \textbf{0.0321} & 0.0627 & 0.0726 & 0.0907 & 0.0520 & 0.0500 & \underline{0.0413} \\
office seq2 part1 & \underline{0.0048} & 0.2062 & 0.2641 & 0.1683 & - & - & \textbf{0.0008} \\
office seq2 part2 & \underline{0.0051} & 0.0820 & 0.1166 & 0.0889 & 0.0610 & - & \textbf{0.0022} \\
office seq3 & 0.0870 & 0.0952 & 0.0328 & 0.0350 & \underline{0.0300} & \underline{0.0300} & \textbf{0.0235} \\
\botrule
\end{tabular}}}
\end{table*} can be calculated:
\begin{equation}
RMSE=(\frac{1}{N}\sum^{N-1}_{i=0}\|trans(\textbf{E}_{i})\|)^{\frac{1}{2}},
\label{E19}
\end{equation}
where $trans(\textbf{E}_{i})$ represents the translation part of the relative pose error. Denote the difference between ground truth and estimated position on $x, y, z$-axis as $\delta x_i, \delta y_i, \delta z_i$, respectively. $trans(\textbf{E}_{i})$ can be expressed by
\begin{equation}
trans(\textbf{E}_{i}) = \|\delta x^2_i + \delta y^2_i + \delta z^2_i\|.
\label{E24}
\end{equation}
In addition, we also use the mean error, median error, and standard derivation error as metrics to evaluate the performance of algorithms.
\begin{tablehere}
\small
\tbl{The accuracy performance on ICL-NUIM Datasets[\citen{icl}]. All results were drawn from the official EVO toolbox[\citen{evo}]. \\ \label{tab1}}
{\resizebox{3.5in}{!}
{
\begin{tabular}{@{}lcccccr@{}}
\toprule
Dataset & RMSE (m) & Mean (m) & Median (m) & Std. (m) & SSE (m)\\ \colrule
office seq0 & 0.021982 & 0.019711 & 0.014278 & 0.009732 & 0.015463\\
office seq1 & 0.041288 & 0.035709 & 0.028995 & 0.020726 & 0.028979 \\
office seq2 part1 & 0.000820 & 0.000794 & 0.000690 & 0.000204 & 0.000002 \\
office seq2 part2 & 0.002198 & 0.001822 & 0.001188 & 0.001230 & 0.000029 \\
office seq3 & 0.023508 & 0.021398 & 0.018970 & 0.009734 & 0.040341 \\
\botrule
\end{tabular}}}
\end{tablehere}
For our self-collected datasets, we employ the final drift error as a performance metric for various algorithms. We utilize Apriltag[\citen{apriltag}] to determine the camera positions for the initial and concluding frames, thereby establishing the ground truth for relative positioning between these frames. Subsequently, we compute the final drift error by comparing this ground truth with the relative positions derived from the algorithm-generated trajectories.

\subsection{Public datasets}
During the experiments, to validate the feasibility of NIDEVO, we used the ICL-NUIM dataset[\citen{icl}] as the benchmark dataset for testing. The ICL-NUIM dataset, introduced by Handa et al. [\citen{icl}] in 2014, is a benchmark dataset for testing RGB-D SLAM performance. This dataset is suitable for our intended application scenario as it provides sequences captured in office environments. We compared the popular algorithms ORB-SLAM2[\citen{orb-slam2}] and RTAB-MAP[\citen{rtab}]. Additionally, we chose OPVO[\citen{OPVO}] and LPVO[\citen{LPVO}], the methods that decouple rotation and translation estimation based on the Manhattan assumption, as a point of comparison. The results of ORB-SLAM2 and RTAB-MAP were reproduced using publicly available code on our own computer, while the results of OPVO and LPVO were cited from the corresponding original paper [\citen{OPVO}] and [\citen{LPVO}]. All algorithms were tested on the same set of image sequences. The experiments were conducted on a 12th Gen Intel® Core™ i5-12500H CPU with 16GB of memory. Our method was implemented using ROS Noetic. The evaluation metrics were calculated using the EVO toolbox[\citen{evo}].

Table~\ref{tab1} shows the accuracy of our NIDEVO on the office scenes dataset from ICL-NUIM. We provided five evaluation metrics to demonstrate our proposed method's performance comprehensively. In the case of the office seq2 sequence, we split it into two segments for evaluation because some images in this sequence do not satisfy the "two planes" assumption that our method relies on. We applied the same approach when comparing with other methods later for fairness.

In Table~\ref{tab2}, we present an exhaustive comparative analysis that includes a range of baseline algorithms. The metrics highlighted in bold within the table are indicative of the best performance achieved on the dataset currently under scrutiny. Upon analyzing the results, it becomes evident that the method we have developed consistently yields performance metrics that are either on par with or superior to those of the baseline algorithms across all the sequences we have tested. This not only highlights the effectiveness and resilience of our suggested approach but also stands as a compelling endorsement of its adaptability to a broader range of application contexts.

The result leads to two important findings. First, it shows that our approach allows for accurate rotation estimation without the need for iterative optimization solvers. Second, our decoupled rotation and translation estimation scheme performs well in terms of accuracy on the ICL-NUIM dataset[\citen{icl}]. These results collectively indicate that our proposed method offers a reliable and efficient alternative for visual odometry tasks.

In our method, the rotation estimation mainly depends on the accuracy of the normal vector map. We apply a mean filtering to process the normal vector map for denoising. To investigate the effect of the filter size, we compare the results with different filter sizes in Table~\ref{tab3_new}. This comparison experiment was conducted on the office seq0 sequence. 

From Table~\ref{tab3_new}, it can be observed that larger values of the cell size led to smaller pose estimation errors. A cell size of 1 indicates no smoothing applied to the normal vectors. However, substantial sliding windows do not necessarily yield better results. Therefore, considering the trade-off between computational efficiency and the local smoothing effect, we ultimately conducted our experiments with a cell size 10.
\begin{tablehere}
\small
\tbl{The influence of mean filter size of the normal map on office seq0. The best results are highlighted in \textbf{bold}.\\ \label{tab3_new}}
{\resizebox{3.5in}{!}{
\begin{tabular}{@{}lcccccr@{}}
\toprule
Cell Size & RMSE (m) & Mean (m) & Median (m) & Std. (m) & SSE (m)\\ \colrule
1 & 0.155239 & 0.141372 & 0.113001 & 0.064135 & 0.819373 \\
5 & 0.061277 & 0.033164 & 0.021836 & 0.051527 & 0.105138 \\
10 & \textbf{0.021982} & \textbf{0.019711} & \textbf{0.014278} & \textbf{0.009732} & \textbf{0.015463} \\
\botrule
\end{tabular}}}
\end{tablehere}
Figure~\ref{office_seq3_traj} offers detailed visualization results of the trajectories generated on the office seq3 of the ICL-NUIM dataset. In this figure, we juxtapose the performance of our proposed method against multiple baseline algorithms as well as the ground truth trajectories for a comprehensive comparison. As we traverse the trajectory, it becomes evident that only a select few methods—namely our own, ORB-SLAM2, and BRISK-RTAB—are capable of closely approximating the ground truth trajectory. What sets our method apart is its superior alignment with the ground truth, particularly in the two regions that are zoomed in for closer examination. This suggests that our algorithm demonstrates a higher degree of accuracy and reliability in these specific areas compared to other competing methods.

In summary, our proposed method generally outperforms other state-of-the-art algorithms, with a notable exception being office seq1 of the ICL-NUIM dataset. In this specific sequence, our approach is marginally outperformed by ORB-SLAM2[\citen{orb-slam2}]. This performance gap is largely due to the fewer overlapping planar points in office seq1, which allows ORB-SLAM2[\citen{orb-slam2}] to leverage sufficient feature point information for robust performance.

On the other hand, in the sequences such as office seq0, office seq2 part1, office seq2 part2, and office seq3 that are low-texture, ORB-SLAM2[\citen{orb-slam2}] performs less optimally compared to our method. Similarly, algorithms like OPVO[\citen{OPVO}] and LPVO[\citen{LPVO}] also fall short in these challenging conditions. The primary reason for their underperformance is their reliance on the Manhattan World assumption, which proves to be less robust in the specific scenarios encountered in these sequences. Moreover, their methods of translation estimation are still dependent on feature points, resulting in diminished accuracy in low-texture settings.

\subsection{Self-collected datasets}
To bolster the empirical validation of our method, we employed datasets self-collected in unique indoor environments. These datasets were procured using a customized ``Magic Helmet'' setup, where we integrated a Realsense L515 camera, a prominent low-cost RGB-D camera. Through this setup, we aimed to simulate real-world AR/VR scenarios, capturing multiple sequences in both office and laboratory settings via handheld operation of the camera. To ensure rigorous algorithm accuracy testing and facilitate comparisons, we strategically positioned \begin{figurehere}
\begin{center}
\centerline{\includegraphics[width=3.5in]{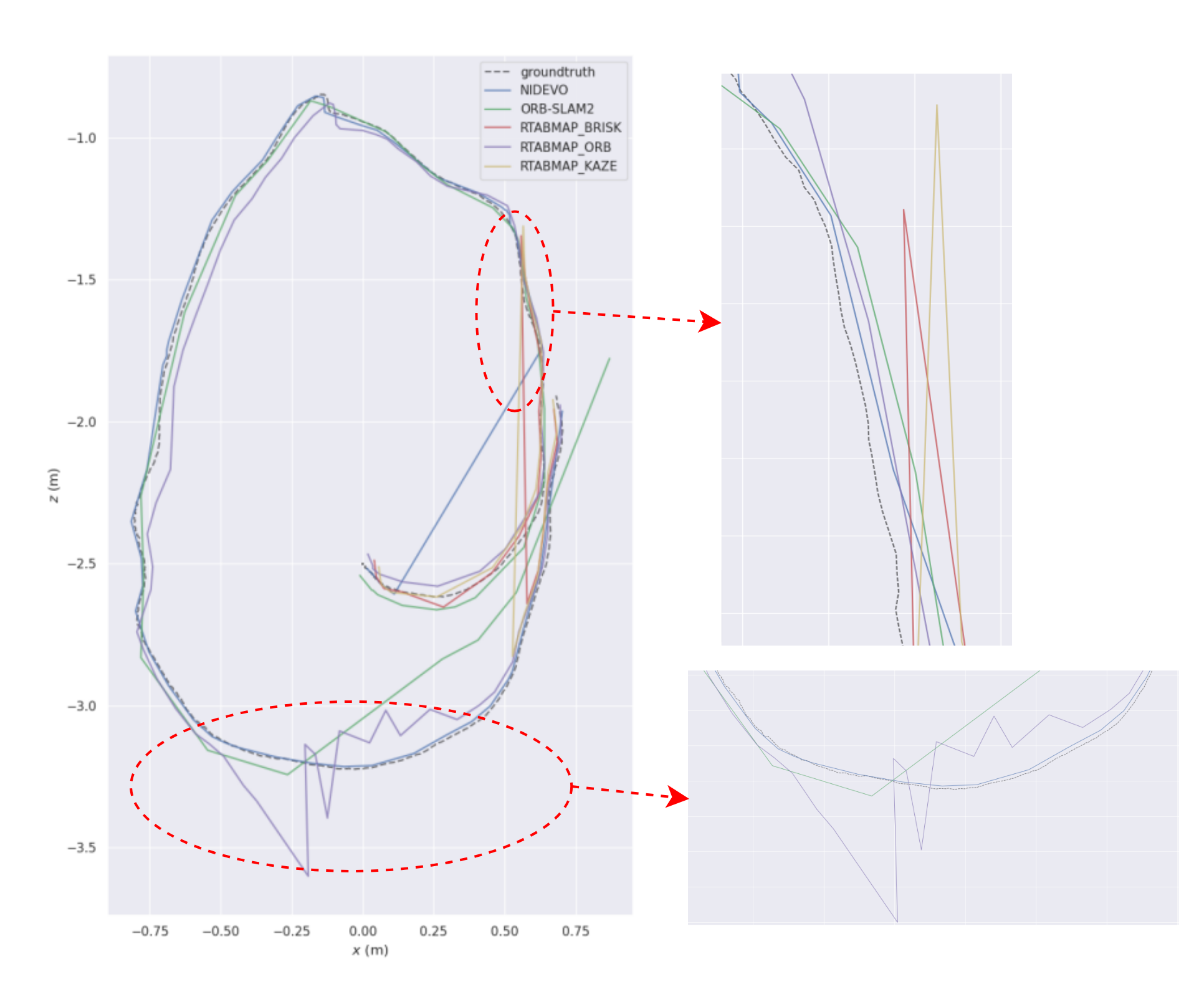}}
\caption{The trajectory comparison on the office seq3. From the zoomed-in views of two local regions, it can be observed that our method (shown in blue) outperforms other methods in terms of performance.}
\label{office_seq3_traj}
\end{center}
\end{figurehere}an AprilTag[\citen{apriltag}] within the recording milieu. This allowed us to capture the same AprilTag[\citen{apriltag}] at both the commencement and conclusion of each sequence, enabling us to compute the camera's pose relative to the AprilTag[\citen{apriltag}] coordinate system at these pivotal junctures. Finally, we can compute the drift error for each algorithm as the Equation~\ref{E23} shown,
\begin{equation}
E = \| ((\textbf{T}^{C0}_{Tag})^{-1} \cdot \textbf{T}^{C1}_{Tag})^{-1} \cdot (\textbf{T}^{E1}_{E0}) \|,
\label{E23}
\end{equation}
where $\textbf{T}$ is the coordinate transform matrix, $_{C0}$ and $_{C1}$ represent the first and last camera frame coordinates, respectively. $Tag$ means the AprilTag[\citen{apriltag}] coordinate. $E0$ and $E1$ represent the first and last estimated camera frame coordinates by algorithm, respectively. 

In our dataset, as illustrated in Figure~\ref{low_texture_scene}, there are low-texture environments accompanied by significant camera jitter during the capture process. We selected two mainstream visual odometry algorithms, ORB-SLAM2[\citen{orb-slam2}] and RTAB-MAP[\citen{rtab}], as our baselines. The comparison results are shown in Table~\ref{tab4}. Our method achieved the best results across all four sequences. Both ORB-SLAM2[\citen{orb-slam2}] and RTAB-MAP[\citen{rtab}] experienced tracking loss in four and two sequences, respectively. In the remaining sequences where ORB-SLAM2[\citen{orb-slam2}] did not suffer from tracking loss, our method accuracy was inferior to that of ORB-SLAM2[\citen{orb-slam2}]. This can be attributed to the limited availability of overlapped planar points in these scenarios, which led to lower accuracy in rotation estimation. However, in summary, our method demonstrates superior robustness compared with both ORB-SLAM2[\citen{orb-slam2}] and RTAB-MAP[\citen{rtab}].
\begin{figurehere}
\begin{center}
\centerline{\includegraphics[width=3in]{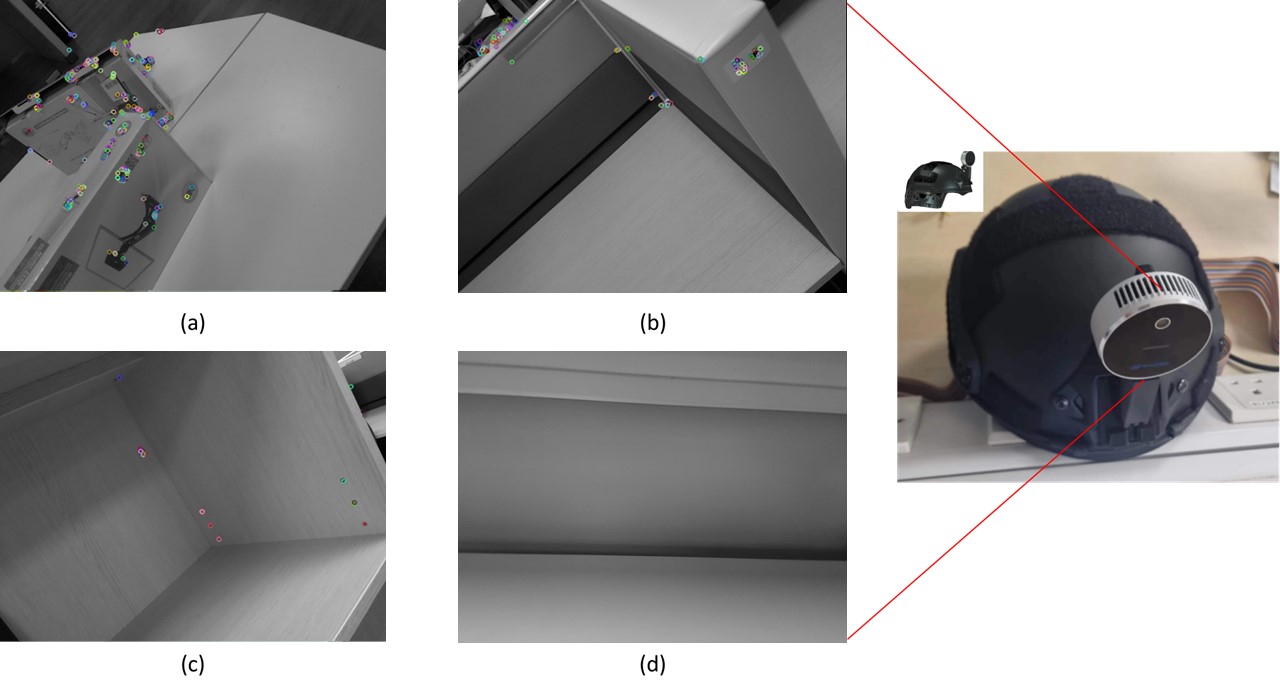}}
\caption{Low-texture sceneries in the self-collected dataset. we try to extract ORB features from these images. It can be seen that there are very few features in (b) and (c). In (d), there are no feature points extracted at all. Although some features are extracted in (a), they are too densely distributed in a small area, which is not ideal for pose estimation.
}
\label{low_texture_scene}
\end{center}
\vspace{-2.5em}
\end{figurehere}
\subsection{Efficiency Analysis}
In this chapter, we analyzed the efficiency of the proposed NIDEVO and discussed the improvements brought by the alternate multithread design. Table~\ref{tab5} shows the processing times of different modules in our system. The rotation estimation module, however, takes the longest time among AP[\citen{rtab}] and experienced tracking loss in four and two sequences, respectively. In the remaining sequences where ORB-SLAM2[\citen{orb-slam2}] did not suffer from tracking loss, the three modules were due to a density-based clustering step. A comparative analysis of the average per-frame processing time between NIDEVO and various baseline algorithms is presented in Table~\ref{tab6}. The computational latency for LPVO is directly sourced from the original publication, whereas the mean processing durations for ORB-SLAM2 and RTAB-MAP are empirically derived through evaluations conducted on the ICL-NUIM dataset. Notably, upon the implementation of parallel computing techniques, NIDEVO emerges as the most computationally efficient solution. Further benchmarking our method on an i9-13900 processor revealed impressive results, with the system achieving speeds of up to 91Hz.
\section{Limitations and Future Work}
The proposed method primarily relies on a single RGB-D sensor, simultaneously capturing both depth and color information. One clear limitation is the dependency on dense depth data. This not only demands significant computational resources but also raises potential concerns regarding the method's resilience to noise. Moving forward, there is significant potential to enhance efficiency by devising an approach that leverages sparse depth information for non-iterative rotation estimation. Another noteworthy aspect is the decoupled nature of our rotation and translation estimation phases, both of which sidestep the need for iterative optimizers. This design ensures minimal computational requirements, making our method particularly suited for deployment on devices with constrained computing capabilities. Envisioning its potential applications, this streamlined computational model holds promise for integration into AR/VR scenarios, where swift and efficient processing is pivotal.

\begin{tablehere}
\vspace{-0.1em}
\tbl{The comparison of drift errors (m) on self-collected datasets. The best results are highlighted in \textbf{bold}. $\times$ refers to tracking-lost. \label{tab4}}
{\resizebox{0.5\textwidth}{!}{
\begin{tabular}{@{}lccccr@{}}
\toprule
Datasets & ORB-SLAM2 & RTAB-MAP & NIDEVO (Ours)\\ \colrule
office table seq1 & $\times$ & 0.815951 & \textbf{0.691397} \\
office table seq2 & \textbf{0.316403} & 0.381820 & 0.805584 \\
office table seq3 & $\times$ & 0.341531 & \textbf{0.158484} \\
cabinet seq1 & $\times$ & $\times$ & \textbf{1.004219} \\
desk seq1 & \textbf{0.210123} & 0.304618 & 0.753913 \\
test-bed seq1 & $\times$ & $\times$ & \textbf{0.297317} \\
\botrule
\end{tabular}}}
\end{tablehere}
\begin{tablehere}
\tbl{The module runtime (ms) for processing a single frame in NIDEVO. \\ \label{tab5}}
{\resizebox{3.5in}{!}{
\begin{tabular}{@{}lcc@{}}
\toprule
& Average Runtime (ms) & Error(ms) \\
\colrule
Normal Map Thread & 7 & $\pm$5 \\
Rotation Thread & 14 & $\pm$5 \\
Translation Thread & 10 & $\pm$5 \\
\botrule
\end{tabular}}}
\end{tablehere}
\begin{tablehere}
\tbl{The efficiency comparison (Hz) of NIDEVO and other baselines. The best result is highlighted in \textbf{bold}. \\ \label{tab6}}
{\resizebox{3.5in}{!}{
\begin{tabular}{@{}cccc@{}}
\toprule
ORB-SLAM2 & RTAB-MAP & LPVO[\citen{LPVO}] & NIDEVO \\ \colrule
59 & 33 & 12.5 & \textbf{71}\\
\botrule
\end{tabular}}}
\end{tablehere}

\section{Conclusions}
In this work, we propose a decoupled VO method based on RGB-D data. In this approach, we first compute the rotation using the Rodrigues formula with a direct method leveraging the structural information in the scene and then estimate the translation using KCC[\citen{kcc}], which is a non-iterative method. 

The rotation estimation primarily relies on the structural information in the depth map, while the translation estimation mainly depends on the frequency information in the orthographic projection map. Hence, our method circumvents the constraints in feature-sparse scenes, thereby achieving more robust tracking performance in low-texture areas. Moreover, thanks to efficient estimation of the normal map, efficient plane clustering in rotation estimation, and efficient solving of phase correlation, our method offers certain advantages in real-time performance compared to mainstream methods. The proposed system has been validated on public datasets and self-collected datasets. The results demonstrate that our system can achieve higher accuracy and efficiency compared to other mainstream RGB-D VO methods.

However, there is still room for improvement in the proposed method. The accuracy of rotation estimation is dependent on the depth map accuracy. For future works, we plan to consider more robust methods for normal map estimation and to consider low-accuracy depth maps as inputs for rotation estimation.

\end{multicols}

\begin{thebibliography}{10}
\bibitem{vinsmono} T. Qin, P. Li, and S. Shen, "Vins-mono: A robust and versatile monocular visual-inertial state estimator," in {\it IEEE Transactions on Robotics}, vol. 34, no. 4, pp. 1004-1020, 2018.

\bibitem{orb-slam2} R. Mur-Artal and J. D. Tardós, Orb-slam2: An open-source slam system for monocular, stereo, and rgb-d cameras, {\it IEEE transactions on robotics} {\bf 33}(5) (2017) 1255--1262.

\bibitem{rtab} M. Labbé and F. Michaud, RTAB‐Map as an open‐source lidar and visual simultaneous localization and mapping library for large‐scale and long‐term online operation, {\it Journal of Field Robotics} {\bf 36}(2) (2019) 416--446.

\bibitem{3rgbd} H. Cui, H. Xue, and H. Hu, "Research on an improved RGB-D camera SLAM odometer algorithm," in {\it International Conference on Computer Graphics, Artificial Intelligence, and Data Processing (ICCAID 2022)}, vol. 12604, pp. 430-436, SPIE, 2023.

\bibitem{4rgbd} A. Fontan, J. Civera, and R. Triebel, "Information-driven direct rgb-d odometry," in {\it Proceedings of the IEEE/CVF Conference on Computer Vision and Pattern Recognition}, 2020, pp. 4929-4937.

\bibitem{rgbd-tam} A. Concha and J. Civera, RGBDTAM: A cost-effective and accurate RGB-D tracking and mapping system, in {\it 2017 IEEE/RSJ international conference on intelligent robots and systems (IROS)} (IEEE, 2017), pp.~6756--6763.

\bibitem{canny-vo} Y. Zhou, H. Li, and L. Kneip, Canny-vo: Visual odometry with rgb-d cameras based on geometric 3-d–2-d edge alignment, {\it IEEE Transactions on Robotics} {\bf 35}(1) (2018) 184--199.

\bibitem{edge-vo} C. Kim, P. Kim, S. Lee, and H. J. Kim, Edge-based robust rgb-d visual odometry using 2-d edge divergence minimization, in {\it 2018 IEEE/RSJ International Conference on Intelligent Robots and Systems (IROS)} (IEEE, 2018), pp.~1--9.

\bibitem{plane-edge-slam} Q. Sun, J. Yuan, X. Zhang, and F. Duan, Plane-Edge-SLAM: Seamless fusion of planes and edges for SLAM in indoor environments, {\it IEEE Transactions on Automation Science and Engineering} {\bf 18}(4) (2020) 2061--2075.

\bibitem{OPVO} P. Kim, B. Coltin, and H. J. Kim, Visual Odometry with Drift-Free Rotation Estimation Using Indoor Scene Regularities, in {\it BMVC}, {\bf 2}(6) (2017), p.~7.

\bibitem{LPVO} P. Kim, B. Coltin, and H. J. Kim, Low-drift visual odometry in structured environments by decoupling rotational and translational motion, in {\it 2018 IEEE international conference on Robotics and automation (ICRA)} (IEEE, 2018), pp.~7247--7253.

\bibitem{Manhattanslam} R. Yunus, Y. Li, and F. Tombari, Manhattanslam: Robust planar tracking and mapping leveraging mixture of manhattan frames, in {\it 2021 IEEE International Conference on Robotics and Automation (ICRA)} (IEEE, 2021), pp.~6687--6693.

\bibitem{kcc} C. Wang, L. Zhang, L. Xie, and J. Yuan, Kernel cross-correlator, in {\it Proceedings of the AAAI Conference on Artificial Intelligence}, {\bf 32}(1) (2018).

\bibitem{ni-slam} C. Wang, J. Yuan and L. Xie, Non-iterative SLAM, in {\it 2017 18th International Conference on Advanced Robotics (ICAR)} (IEEE, 2017), pp.~83--90.

\bibitem{rgbd-dso} Z. Yuan, K. Cheng, J. Tang, and X. Yang, Rgb-d dso: Direct sparse odometry with rgb-d cameras for indoor scenes, {\it IEEE Transactions on Multimedia} {\bf 24} (2021) 4092--4101.

\bibitem{liu-manhattan} J. Liu and Z. Meng, "Visual slam with drift-free rotation estimation in manhattan world," {\it IEEE Robotics and Automation Letters}, vol. 5, no. 4, 2020, pp. 6512-6519.

\bibitem{rgbd-regular} Y. Li, R. Yunus, N. Brasch, N. Navab, and F. Tombari, "RGB-D SLAM with structural regularities," in {\it 2021 IEEE International Conference on Robotics and Automation (ICRA)}, pp. 11581-11587, IEEE, 2021.

\bibitem{linear-rgbd} K. Joo, P. Kim, M. Hebert, I. S. Kweon, and H. J. Kim, "Linear RGB-D SLAM for structured environments," {\it IEEE Transactions on Pattern Analysis and Machine Intelligence}, vol. 44, no. 11, 2021, pp. 8403-8419.

\bibitem{linear-atlanta} K. Joo, T.-H. Oh, F. Rameau, J.-C. Bazin, and I. S. Kweon, "Linear rgb-d slam for atlanta world," in {\it 2020 IEEE International Conference on Robotics and Automation (ICRA)}, pp. 1077-1083, IEEE, 2020.

\bibitem{atlanta} K. Joo et al., "Globally optimal inlier set maximization for Atlanta world understanding," {\it IEEE Transactions on Pattern Analysis and Machine Intelligence}, vol. 42, no. 10, 2019, pp. 2656-2669.

\bibitem{vote} R. Martins, E. Fernandez-Moral, and P. Rives, An efficient rotation and translation decoupled initialization from a large field of view depth images, in {\it 2017 IEEE/RSJ International Conference on Intelligent Robots and Systems (IROS)} (IEEE, 2017), pp.~5750--5755.

\bibitem{two-plane} J.-C. Bazin, C. Demonceaux, P. Vasseur, and I. S. Kweon, Motion estimation by decoupling rotation and translation in catadioptric vision, {\it Computer Vision and Image Understanding} {\bf 114}(2) (2010) 254--273.

\bibitem{icl} A. Handa, T. Whelan, J. McDonald, and A. J. Davison, A benchmark for RGB-D visual odometry, 3D reconstruction and SLAM, in {\it 2014 IEEE international conference on Robotics and automation (ICRA)} (IEEE, 2014), pp.~1524--1531.

\bibitem{apriltag} J. Wang and E. Olson, AprilTag 2: Efficient and robust fiducial detection, in {\it 2016 IEEE/RSJ International Conference on Intelligent Robots and Systems (IROS)} (IEEE, 2016), pp.~4193--4198.

\bibitem{evo} M. Grupp, evo: Python package for the evaluation of odometry and SLAM (2017).
\end{thebibliography}
\end{document}